\documentclass[11pt,letterpaper]{article}

\usepackage{tabularx,booktabs}
\usepackage{url}
\usepackage{fullpage}
\usepackage{graphicx}
\usepackage{lipsum}
\usepackage{booktabs} 
\usepackage{algorithm}
\usepackage{algorithmic}
\usepackage{amsmath}
\usepackage{multirow, makecell}
\usepackage{url}
\usepackage{flushend}
\usepackage{comment}
\usepackage{xspace}
\usepackage{adjustbox}
\usepackage{color}
\usepackage{enumerate}
\usepackage{enumitem}
\usepackage{caption}
\usepackage[margin=1in]{geometry}
\usepackage[hang,flushmargin]{footmisc} 

\newcommand{\junk}[1]{}
\newcounter{rownumbers}

\newcommand{\name}{{\sc Athena}\xspace}

\newcommand{\ie}{\emph{i.e.,}\xspace}
\newcommand{\eg}{\emph{e.g.,}\xspace}

\definecolor{ballblue}{rgb}{0.13, 0.67, 0.8}

\DeclareMathOperator*{\argmin}{argmin}

\begin{document}
\thispagestyle{empty}

\title{\textit{\name}: Automated Tuning of Genomic Error Correction Algorithms using Language Models}

\author
{
\centering
Mustafa Abdallah$^{*1}$  \and Ashraf Mahgoub$^{*2}$ 
\and Saurabh Bagchi$^{\dagger 1,2}$ \and Somali Chaterji$^{1}$ 
\\
$^{1}$School of Electrical and Computer Engineering\\
$^{2}$Department of Computer Science\\
Purdue University\\
{\small $^*$ The first two authors contributed equally to the work.} \\
{\small $\dagger$ Contact author.}
}

\date{}

\maketitle
\begin{abstract}
  \noindent {The performance of most error-correction algorithms that operate on genomic sequencer reads is dependent on the proper choice of its configuration parameters, such as the value of $k$ in \textit{k}-mer based techniques.} In this work, we target the problem of finding the best values of these configuration parameters to optimize error correction.
We perform this in a data-driven manner, due to the observation that different configuration parameters are optimal for different datasets, \ie from different instruments and organisms. 
We use language modeling techniques from the Natural Language Processing (NLP) domain in our algorithmic suite, \textsc{Athena}, to automatically tune the performance-sensitive configuration parameters. Through the use of N-Gram and Recurrent Neural Network (RNN) language modeling, we validate the intuition that the EC performance can be computed quantitatively and efficiently using the ``perplexity'' metric, prevalent in NLP. After training the language model, we show that the perplexity metric calculated for runtime data has a strong negative correlation with the correction of the erroneous NGS reads. Therefore, we use the perplexity metric to guide a hill climbing-based search, converging toward the best $k$-value. Our approach is suitable for both \textit{de novo} and comparative sequencing (resequencing), eliminating the need for a reference genome to serve as the ground truth. This is important because the use of a reference genome often carries forward the biases along the stages of the pipeline. 

\end{abstract}

\section{Introduction}
\begin{figure}
\centering
\includegraphics[width=\linewidth]{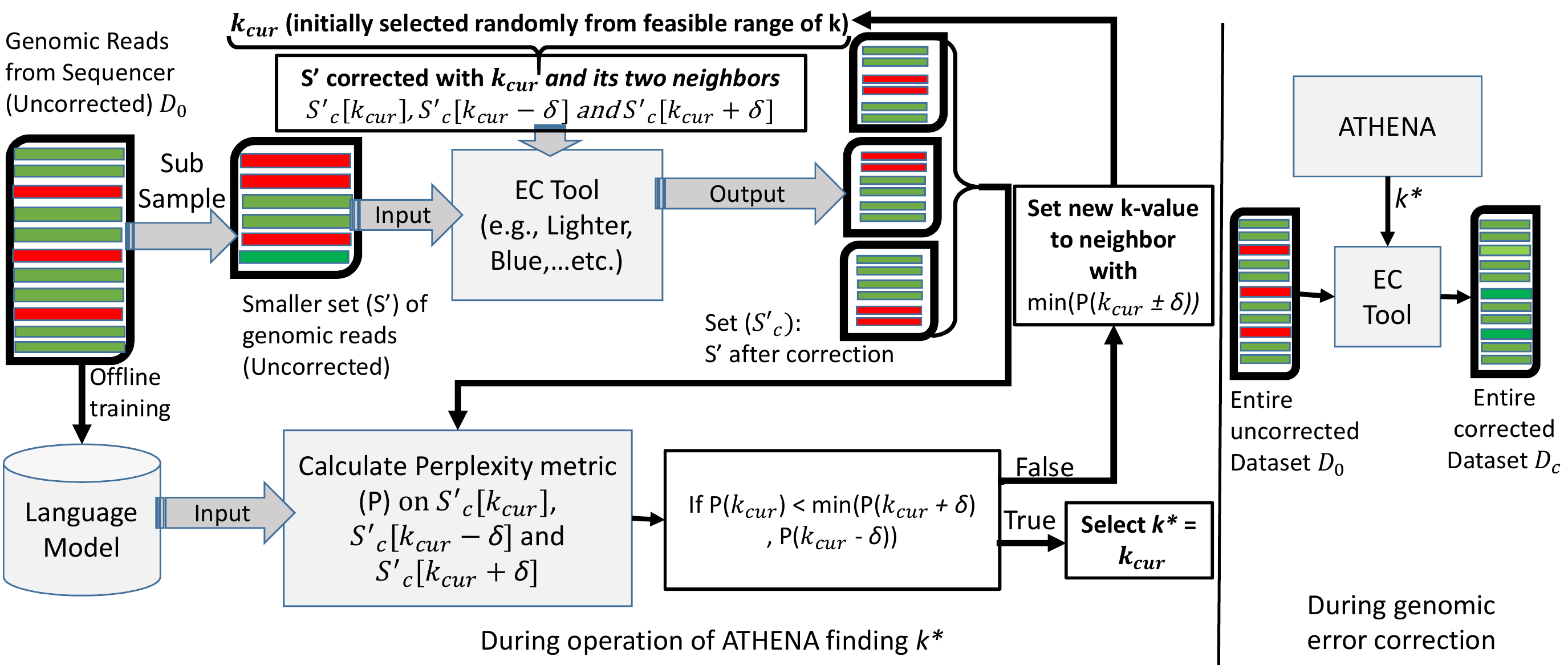}
\caption{Overview of \name's workflow. First, we train the language model using the entire set of uncorrected reads for the specific dataset. Second, we perform error correction on a subsample from the uncorrected reads using an EC tool (\eg Lighter or Blue) and a range of $k$-values. Third, we compute perplexity of each corrected sample, corrected with a specific EC tool, and decide on the best $k$-value for the next iteration, \ie the one corresponding to the lowest perplexity metric because EC quality is anti-correlated with the perplexity metric. This process continues until the termination criteria are met. Finally, the complete set of reads is corrected with the best $k$-value found and then used for evaluation.}	
\label{fig:AthenaOverview}
\end{figure}
\vspace{-10pt}
Rapid advances in next-generation sequencing (NGS) technologies, with the resulting drops in sequencing costs, offer unprecedented opportunities to characterize genomes across the tree of life. 
While NGS techniques allow for rapid parallel sequencing, they are more error-prone than Sanger reads 
and generate different error profiles, \eg substitutions, insertions, and deletions. 
The genome analysis workflow needs to be carefully orchestrated so errors in reads are not magnified downstream.
Consequently, multiple error-correction (EC) techniques have been developed for improved performance for applications ranging from \textit{de novo} variant calling to differential expression, iterative $k$-mer selection for improved genome assembly \cite{mahadik2017scalable} and for long-read error correction \cite{szalay2015novo, sameith2016iterative}. 

Importantly, the values of the performance-sensitive configuration parameters are dependent not only on the dataset but also on the specific EC tool (Table \ref{tb1:Lighter-Blue-Racer-Perplexity-vs-Alignment}, Table \ref{tb1:Lighter-Perplexity-vs-Alignment} in Appendix). The performance of many EC algorithms is highly dependent on the proper choice of configuration parameters, \eg $k$-value (length of the substring) in \textit{k}-spectrum-based techniques.  
Selecting different $k$-values has a trade-off such that small values increase the overlap probability between reads, however, an unsuitably small value degrades EC performance because it does not allow the algorithm to discern correct $k$-mers from erroneous ones. In contrast, unsuitably high $k$-values decrease the overlap probability and hurts EC performance because most $k$-mers will now appear unique. The $k$-mers that appear above a certain threshold frequency, and are therefore expected to be legitimate, are \textit{solid k-mers}, the others are called \textit{insolid or untrusted $k$-mers}. In $k$-spectrum-based methods, the goal is to convert insolid $k$-mers to solid ones with a minimum number of edit operations. 
Thus, an adaptive  method for finding the best $k$-value and other parameters is needed for improved EC performance, and in turn, genome assembly. 

Many existing EC solutions (\eg Reptile \cite{yang2010reptile}, Quake \cite{kelley2010quake}, Lighter \cite{song2014lighter}, Blue \cite{greenfield2014blue}) require users to specify $k$. The best value is usually found by exploration over $k$-values \cite{kao2011echo} and evaluating performance metrics, \eg EC Gain, Alignment Rate, Accuracy, Recall, and Precision.
However, a reference genome is typically needed to serve as ground truth for such evaluations, making this approach infeasible for \textit{de novo} sequencing tasks. 
To the best of our knowledge, existing tools leave the best parameter choice to the end user~\cite{peng2010idba, mahadik2017scalable}, although \textsc{Kmergenie} provides intuitive abundance histograms or heuristics to guide $k$-value selection, albeit for genome assembly. 
Further, \textsc{Kmergenie} does not provide the optimal $k$-value for different tools, only accounting for the dataset for computing the abundance histograms. We find that this is an incorrect simplification, \eg the optimal $k$-values for Blue and Lighter in our evaluation vary (Tables  \ref{tb1:Lighter-Blue-Racer-Perplexity-vs-Alignment}, \ref{tb1:Lighter-Perplexity-vs-Alignment}, and \ref{tb1:Blue-Perplexity-vs-Alignment}). 
Within assembly also, it is well known that different $k$-values are optimal for different assembly pipelines \cite{Limasset2017TowardPR}.
Thus, the need for a data-driven method to select the optimal $k$-value is crucial. 
For example, we found that $k$=25 gives the best EC gain performance for D5 using Lighter. However, if the same $k$-value was used for D3, the EC gain is only 65\% compared to the gain of 95.34\% when using the optimal $k$-value (Table \ref{tb1:Lighter-Perplexity-vs-Alignment} in Appendix).

\begin{figure*}
  \includegraphics[width=\linewidth]{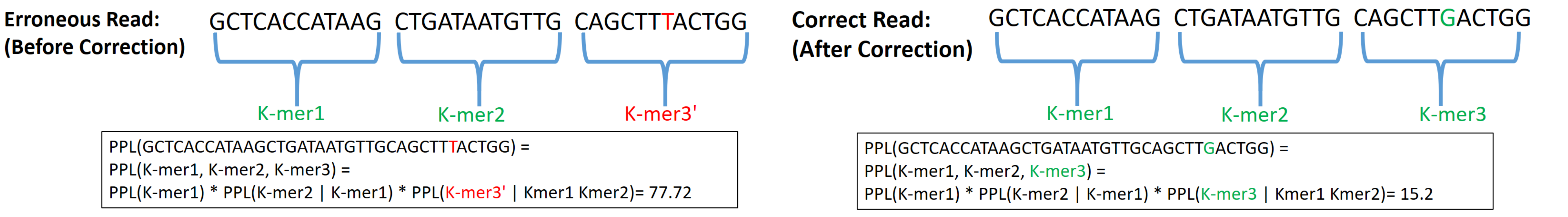}
  \caption{An example showing how the perplexity metric encodes errors in genomic reads. The read on the left is an erroneous read selected from dataset \#3 (D3), while the read on the right is the same read, after correction with Lighter. When using language modeling to compute the perplexity for both reads, we notice that the read on the right has a lower perplexity value (15.2), relative to the erroneous read (77.72), as the sequence of $k$-mers after correction has a higher probability of occurrence. Also notice that the probability of a sequence of $k$-mers depends on both their frequencies and their relative order in the read, which allows the perplexity metric to capture how likely it is to observe this $k$-mer given the neighboring $k$-mers in a given read.}
  \label{fig:PPL_Simple_Example}
\end{figure*}

Our solution, {\em \name}\footnote{Just as \textit{Athena} is the Greek Goddess of wisdom and a fierce warrior, we wish our technique to unearth the genomic codes underlying disease in a fearless war against maladies.}, finds the best value of the configuration parameters
for correcting errors in genome sequencing, such as the value of $k$ in $k$-mer based methods. Further, \name does not require access to a reference genome to perform its function of determining the optimal parameter configuration\footnote{In our evaluation, we use Bowtie2 for alignment and measure alignment rate as a metric to evaluate \name. However, alignment is \textit{not} needed for \name to work.}. 
\name, like other EC tools, leverages the fact that NGS reads have the property of reasonably high coverage, 30X--150X coverage depth is commonplace. 
From this, it follows that the likelihood of correct overlaps for a given portion of the genome will outnumber the likelihood of erroneous ones \cite{yang2010reptile}.
\name captures the underlying semantics of NGS reads from different datasets such that the language model (LM) can learn the complexities of a language and can also predict future sequences. 
This is integral to traditional NLP tasks such as speech recognition, machine translation, or text summarization. LM's goal is to learn a probability distribution over sequences of symbols and words,
in this case specialized to the dataset \textit{and} to the EC tool of choice for optimal performance.
Therefore, it can be used to identify sequences that do not fit the semantics of the language and thus contribute to a high value of the {\em ``Perplexity metric''}. This personalization is conducive to our problem because the LM then learns the patterns pertaining to the specific genome.

In our context, we use LM to estimate the probabilistic likelihood that some observed sequence is solid or insolid, in the context of the specific genome. We train an LM using the original dataset (with uncorrected reads), and then use this trained model to compute the Perplexity metric when the EC tool is run with a specific configuration, \ie for specific parameters. 
We show empirically that the perplexity metric is inversely correlated to the quality of error correction.
Crucially, the perplexity metric evaluation does not require the computationally expensive alignment to a reference genome.
Through a stochastic optimization method, we evaluate the search space to pick the best configuration parameter value to guide EC---$k$ in $k$-mer-based methods and the Genome Length (GL) in the RACER EC tool. 
Moreover, most EC tools are evaluated based on their direct ability to reduce error rates rather than to improve genome assembly. Although the assembly can benefit substantially from EC tools, this can also sometime negatively affect assembly quality due to conversion of benign errors into damaging errors \cite{heydari2017evaluation}. In our evaluation, we see that the EC improvement due to \name also leads to higher quality assembly (Table \ref{tbl:Fiona_5_datasets_Correlation_N_Gram}).

\noindent{In summary, this paper makes the following contributions}. 
\vspace{-6pt}
\begin{enumerate}
\item \name develops a Language Model-based suite to efficiently determine the best configuration parameter value for any existing Error Correction tool. 
\vspace{-8pt}
\item We introduce a likelihood-based metric, the Perplexity metric repurposed from NLP, to evaluate EC quality that does not need a gold standard genome. This is the first such use of this metric in the bioinformatics domain.
\vspace{-8pt}

\item We compare and contrast two LM variants of \name, N-gram and RNN-based. Through this, we show that N-Gram modeling can be faster to train while char-RNN provides similar accuracy to N-gram, albeit with significantly lower memory footprint, 
conducive to multi-tenant analysis pipelines \eg MG-RAST for metagenomics processing \cite{meyer2017mg}.
\vspace{-8pt}

\item We apply \name to 3 EC tools: Lighter \cite{song2014lighter}, Blue \cite{greenfield2014blue}, and RACER \cite{ilie2013racer} on 5 real datasets, with varied error rates and read lengths. We show that \name was successful in finding either the best parameters ($k$ for Lighter and Blue, and $Genome Length$ for RACER) or parameters that perform within 0.27\% in overall alignment rate to the best values using exhaustive search against a reference genome. We couple EC, with best parameter selection by \name, to the Velvet assembler and find that it improves assembly quality (NG50) by a Geometric Mean of 10.3X across the 5 datasets. 
\end{enumerate}


\section{Our Solution: {\bf \name}}
\subsection{Application of Language Models}
We use two different LMs in \name, which gives two different variants. We describe them next.

\noindent \textbf{N-Gram Language Models}: 
We train an N-Gram model \cite{brown1992class}, which is word-based, from the input set of reads before correction. This N-Gram model needs word-based segmentation of the input read as a pre-processing phase. Then, we use this trained LM to evaluate EC performance.

\noindent \textbf{RNN Language Models}: The second technique is RNN-based LM \cite{kombrink2011recurrent}, using different RNN architectures, \eg standard RNNs, LSTMs, and GRUs.
These models can be trained either as word-based models or character-based models. We train our RNN variant as character-based models to avoid having to make the decision about how to segment the genomic string, as we have to do for the N-Gram model.


\noindent{Contrasting our 2 LM variants:} Although N-Gram LMs are much faster compared to RNN-based models, they still have the requirement of splitting a read into words of specific length.
Further, RNN-based models have much lower memory footprint and storage requirements relative to N-Gram. This is because N-Gram models need to store conditional probabilities in large tables with an average size of 0.6--1.1 GB across the 5 datasets. In contrast, RNNs only need to store the network architecture and  weights with an average size of 3--5 MB across the 5 datasets). 

\subsection{Intuition for the Use of the Perplexity metric}
Our design uses the Perplexity metric to provide an accurate, and importantly quick, estimation of the EC quality with the current configuration parameter value(s). 
The Perplexity metric is based on using LM trained on the entire original data prior to error correction.
The Perplexity metric is then calculated on a subset of the corrected reads to measure the  EC performance. It measures how well language modeling can predict the next element in an input stream. Suppose the input stream is $H$ and the next element is $e$. Then, the Perplexity metric is inversely proportional to the probability of seeing ``e'' in the stream, given the history $H$ for the learned model.
This method works because there is a high negative correlation of the Perplexity metric with both EC metrics---Alignment Rate and EC Gain.
Given this anti-correlation, we can rely on the Perplexity metric as an evaluation function, and apply a simple search technique (\textit{e.g.}, hill climbing) to find the best $k$-value for a given dataset. In this description, for simplicity of exposition, we use the $k$-value in $k$-mer based techniques as an example of \name-tuned configuration parameter. However, \name can tune any other relevant configuration parameter in error correction algorithms and we experimentally show the behavior with another parameter---Genome Length---in RACER. 
Figure \ref{fig:PPL_Simple_Example} shows an example how Perplexity can evaluate the likelihood of a sequence of $k$-mers using their frequencies and contextual dependencies. In this example, we notice that the corrected read set (\ie on the right) has a considerably lower Perplexity value (15.2), relative to the erroneous set (77.72). Thus, our intuition that the Perplexity metric reflects the correctness of the read dataset is validated through the observed negative relationship.
For an N-Gram language model, perplexity of a sentence is the inverse probability of
the test set, normalized by the number of words. It is clear from \eqref{eq:4} that minimizing perplexity is the same as maximizing the probability of the observed set of $m$ words from $W_{1}$ to $W_{m}$. For RNN, perplexity is measured as the exponential of the mean of the cross-entropy loss (CE) as shown in \eqref{eq:5}, where $\hat{y}$ is the predicted next character---the output of the RNN---and $|V|$ is the vocabulary size used during training.
 
\begin{equation}\label{eq:4}
\footnotesize       
PP(W) = \sqrt[m]{\frac{1}{P(W_{1},W_{2},....,W_{m})}} \approx \sqrt[m]{\frac{1}{\displaystyle \prod_{i=1}^{m}P(W_{i} | W_{i-1},...,W_{i-n})}}
\end{equation}

\begin{equation}\label{eq:5}
\footnotesize       
CE(y, \hat{y}) = - \sum_{i=1}^{|V|} p(y_{i})\log(p(\hat{y_i})).
\end{equation}

\begin{minipage}[t]{.4\textwidth}
\footnotesize       
\begin{equation}\label{eq:perplexity_fun}
\text{Perplexity} = f(LM, D_C, k)
\end{equation} 
\end{minipage}
\hfill
\begin{minipage}[t]{.4\textwidth}
\footnotesize       
\begin{equation}\label{eq:hill_climb}
  k_{opt} = \argmin_{k_i} f(LM, D_C, k_i)
\end{equation}
\end{minipage}


\subsection{Search through the Parameter Space}
Our objective is to find the best $k$-value that will minimize the Perplexity of the corrected dataset. The Perplexity function is denoted by $f$ in \eqref{eq:perplexity_fun}. Here, LM: trained language model, $D_C$: corrected genomic dataset, and $k$: the configuration parameter we wish to tune.
$f$ is a discrete function as $k$-values are discrete, and therefore, its derivative is not computable. Thus, a gradient-based optimization technique will not work. Hence, we use a simple hill-climbing technique to find the value of $k$ that gives the minimum value of $f$, for the given LM and $D_C$ \eqref{eq:hill_climb}.

\noindent{The following pseudo-code describes the steps used for finding the best $k$-value for a given dataset.} We start with Algorithm 1, which invokes Algorithm 2 multiple times, each time with a different starting value. We begin by training an LM on the original uncorrected read set ($D_0$). Second, we assume that the best value of $k$ lies in a range from $A$ to $B$ (initially set to either the tool's recommended range, or between 1 and $L$, where $L$ is the read size).

The following pseudo-code describes the steps used for finding the best $k$-value for a given dataset. We start with Algorithm 1, which invokes Algorithm 2 multiple times, each time with a different starting value. We begin by training a language model on the original uncorrected read set ($D_0$). Second, we assume that the best value of $k$ lies in a range from $A$ to $B$ (initially set to either the tool's recommended range, or between 1 and $L$, where $L$ is the read size).
We apply an existing EC algorithm (Lighter, Blue, or RACER in our evaluation) with different initial values $(k_{0}$,..., $k_{m}) \in (A, B)$, to avoid getting stuck in a local minimum, going through multiple iterations for a given initial value. We evaluate the Perplexity for the corrected dataset with current value of $k$, $k_i$, and its neighbors ($k_i-\delta$ and $k_i+\delta$). In each iteration, we apply hill-climbing search to identify the next best value of $k_{i}$ for the following iteration. The algorithm terminates whenever the Perplexity relative to $k_i$ is less than the perplexities of both its neighbors or the maximum number of (user-defined) iterations is reached. However, all shown results are with respect to only one initial value (\textit{i.e.}, m=1 in k1, k2, ..., km).


\begin{minipage}[t]{.43\textwidth}
\begin{algorithm}[H]
\caption{Correct Set of Reads}
\begin{algorithmic} 
\scriptsize
\REQUIRE Dataset: $D_{0}$, Read Length: $L$, Maximum number of iterations: $IterMax$ \\
\ENSURE Corrected\_Dataset: $D_c$ 
\begin{flushleft}
\vspace{-7pt}
\textbf{1-} Train a Language Model (LM) using $D_{0}$. \\
\textbf{2-} Select random sample $S'$ from $D_{0}$.\\
\textbf{3-} Pick $k_0$,$k_1$,...,$k_m$: initial values of $k$ in the range of (1, L).\\
\textbf{4-} for i in the range (0, m)\\ 
\textbf{5-} \quad Call \textbf{Find Optimal $K_i$}($S'$, L, LM, IterMax, $k_i$)\\
\textbf{6-} Take $k$ as the argmin of $K_i$ values picked at step 5 and do a complete correction on the entire dataset $D_0$ and return $D_c$.
\end{flushleft}
\end{algorithmic}
\end{algorithm}
\end{minipage}
\begin{minipage}[t]{.52\textwidth}
\scriptsize
\begin{algorithm} [H]
\caption{Find Optimal $k$}
\begin{algorithmic} 
\scriptsize
\REQUIRE Data set: $S'$, Read Length: L, Language Model: LM, MaxNumOfIterations: IterMax, Initial value of $k$: $k_i$ 
\ENSURE $Best k: k*$
\begin{flushleft}
\vspace{-7pt}
\textbf{1-} Evaluate Perplexity of $k_i$ and its neighbors $(k_i-\delta, k_i+\delta)$.\\
\textbf{2-} If $f(LM,S',k_i) < f(LM,S',k_i-\delta) \&\& f(LM,S',k_i) < f(LM,S',k_i+\delta)$, return $k_i$ as $k*$ \\
\textbf{3-} Else, set $k_{i+1}$ = value of neighbor with lower $f$ value.\\
\textbf{4-} if IterMax $>$ 0 Do:\\
\quad \textbf{5-} decrement IterMax and Call \textbf{Find Optimal K} (S', L, LM, IterMax , $k_{i+1}$)\\
\textbf{6-} else \\
\textbf{7-} \quad return best $k_i$ so far
\newline
\end{flushleft}
\end{algorithmic}
\end{algorithm}

\end{minipage}

\subsection{Time and Space Complexity}
Because we apply hill climbing search to find the best value of $k$, the worst-case time complexity of the proposed algorithm is $L$, the upper bound of the range of $k$ values to search for. For the space complexity, \name only needs to save the Perplexity values of previously investigated values of $k$, which is also linear in terms of $L$.


\vspace{-5pt}
\section{Evaluation with Real Datasets}
\label{sec:evaluation-real}

In this section, we evaluate \name variants separately by correcting errors in 5 real datasets and evaluating the quality of the resultant assembly. We implement the N-Gram model using the SRILM toolkit~\cite{Stolcke02srilm--}. SRILM is an open source toolkit that supports building statistical N-Gram LMs, in addition to evaluating the likelihood of new sequences using the perplexity metric. For the RNN LM implementation, we build on the TensorFlow platform \cite{45381}.
After correction, we run the Bowtie2 aligner \cite{langmead2012fast} and measure the Alignment Rate and the Error Correction Gain. A higher value for either metric (i.e., Alignment Rate or Correction Gain) implies superior error correction.
We do a sweep through a range of $k$-values and measure the alignment rate to determine if the \name-generated $k$-value is optimal or its distance from optimality. For interpreting the execution time results, our experiments were performed on Dell Precision T3500 Workstation, with 8 CPU cores, each running at 3.2GHZ, 12GB RAM, and Ubuntu 16.04 Operating System.
We use 3 EC tools, in pipeline mode with \name, namely, Lighter, Blue, and RACER. Blue uses a k-mer consensus to target different kinds of errors such as substitution, deletion and insertion errors, as well as uncalled bases. This improves the performance of both alignment and assembly \cite{greenfield2014blue}. On the other hand, Lighter is much faster as it uses only a sample of k-mers to perform correction. 
Third, RACER uses a different configuration parameter than the value of $k$ and we are able to tune that as well. Specifically, RACER uses the genome length to automatically calculate multiple $k$-values to use for correction.
Our ability to tune any of these EC algorithm's parameters is in line with our vision and ongoing work to design extensible blocks of software to expedite algorithmic development in bioinformatics~\cite{mahadik2016sarvavid}.
Incidentally, we started using another popular EC tool, Reptile, but it only allowed for a smaller range of $k$-values, beyond which it ran into out-of-memory errors. Hence, to demonstrate results with the full range of $k$ values, we restricted our pipelining to Lighter and Blue. 

Our datasets are Illumina short reads [Table \ref{tbl:Datasets_Description}], used in multiple prior studies (\eg~\cite{yang2010reptile, doi:10.1093/bioinformatics/btp379}). For these, there exist ground-truth reference genomes, which we use to evaluate the EC quality. 
The five datasets have different read lengths (from 36bp in D1\&D3 to 100pb in D5) and different error rates (from $<$ 3\% in D1 to 43\% in D2).  

\begin{table}[H]
\begin{center}
\caption {Datasets' description with coverage, number of reads, length of each read, and the reference genome.} 
\label{tbl:Datasets_Description}
\small 
\scalebox{0.86}{
 \begin{tabular}{|c|c|c|c|c|c|c|c|} 
 \hline
 Dataset &  Coverage & \#Reads & Read Length & Genome Type & Reference Genome & Accession Number \\ [0.5ex] 
 \hline
 \multirowcell{1}{D1} & 80X & 20.8M &  \multirowcell{1}{36 bp} & \multirowcell{1}{\textit{E. coli} str. K-12 substr} & NC\_000913  & SRR001665 \\
 \hline
 D2 & 71X & 7.1M & 47 bp & \multirowcell{1}{\textit{E. coli} str. K-12 substr} & NC\_000913 & SRR022918  \\ 
 \hline
 D3 & 173X & 18.1M & 36 bp & \multirowcell{1}{\textit{Acinetobacter} sp. ADP1} & NC\_005966 & SRR006332   \\ 
  \hline
  D4 & 62X & 3.5M & 75 bp & \multirowcell{1}{\textit{B. subtilis}} & NC\_000964.3 & DRR000852  \\
   \hline
  D5 & 166X & 7.1M & 100 bp & \multirowcell{1}{\textit{L. interrogans C} sp. ADP1} & NC\_005823.1 & SRR397962    \\
 \hline
\end{tabular}
}
\end{center}
\end{table}

\vspace{-12pt}
\subsection{Optimal Parameter Selection}
\vspace{-8pt}
The results of using \name with Lighter, Blue, and RACER tools are shown in Table \ref{tb1:Lighter-Blue-Racer-Perplexity-vs-Alignment} for each dataset. In the third column, the value of $k$ (or, GL for RACER) in bold is the one that gives the best assembly quality, evaluated using ground truth.
We see that this almost always corresponds to the lowest perplexity scores computed using \name's language modeling (Table \ref{tb1:Lighter-Perplexity-vs-Alignment}, Table \ref{tb1:Blue-Perplexity-vs-Alignment}). 
In the cases where the lowest perplexity does not correspond to the best assembly quality (this happens in 3 of the 10 cases aggregated between Lighter and Blue), the difference in overall alignment rate is small, less than 0.27\% in the worst case.  
Further, the anti-correlation between perplexity and alignment rate holds for both optimal and non-optimal $k$-values. 
This shows that our hypothesis is valid across a range of $k$-values. We notice that the feasible range of $k$-values in Blue is (20, 32), distinct from Lighter's. Another interesting observation is that the optimal $k$-values are different across the two different EC tools, Lighter and Blue, for the same dataset, as observed before~\cite{song2014lighter}). 
\name can be applied to a different configuration parameter, GL for the RACER tool, in line with our design as a general-purpose tuning tool.

\vspace{-15pt}
\subsection{N-Gram Language Model Results}
\vspace{-5pt}
We start by training an N-Gram language model from the original dataset. We divide each read into smaller segments (words) of length $L_s$ (set to 7 by default). A careful reader may be concerned that selecting the best value of $L_s$ is a difficult problem in itself, of the same order of difficulty as our original problem. Fortunately, this is {\em not} the case and $L_s$ selection turns out to be a much simpler task. From domain knowledge, 
we know that if we use $L_s = 4$ or less, the frequencies of the different words will be similar,
thus reducing the model's discriminatory power, while a large value will mean that the model's memory footprint increases. We find that for a standard desktop-class machine with 32GB of memory, $L_s = 8$ is the maximum that can be accommodated. Further, we find that the model performance is {\em not} very sensitive in the range (5--7), so we end up using $L_s = 7$.
The same argument holds for selecting a history of words, and we use a tri-gram model (history of 3 words) for all our experiments.
Second, we compare the perplexity metric for datasets corrected with different $k$-values and compare the perplexity metric (without a reference genome) to the alignment rate (using a reference genome). 
We always report the \textit{average perplexity}, which is just the total perplexity averaged across all words.
Our results show a high negative correlation ($\leq$ -0.946) between the two  metrics on the 5 datasets, as shown in Table \ref{tbl:Fiona_5_datasets_Correlation_N_Gram}. 
To reiterate, the benefit of using the perplexity metric is that it can be computed without the ground truth and even where a reference genome is available, it is more efficient than aligning to the reference and then computing the alignment rate. 

\begin{table}
\centering
 \caption {Comparison of Lighter, Blue, and RACER using 5 datasets. This is for finding the best $k$-value (GL for RACER) using \name variants \textit{vs.} exhaustive search. We find either the optimal value or within 0.27\% (over Alignment Rate) and within 8.5\% (EC Gain) of the theoretical best (in the worst case), consistent with the reported results by Lighter (Figure 5 in \cite{song2014lighter}). We also notice that for RACER, GL found by \name is within 3\% of the reference GL (except for the RNN model with D5, which still achieves very close performance for both Alignment Rate and EC Gain.)} 
\scalebox{0.75}{
\small
\begin{tabular}{ |c|c|c|c|c|c|c|c|c|c| } 
\hline
\multicolumn{10}{|c|}{\textbf{Lighter}}\\
\hline
{\textbf{Dataset}} & \multicolumn{3}{c|}{\textbf{With \name (N-gram)}}& \multicolumn{3}{c|}{\textbf{With \name (RNN)}} & \multicolumn{3}{c|}{\makecell{\textbf{Exhaustive Search}}} \\ 
 \hline
 					 & \makecell{\textbf{Selected}\\ \textbf{$k$}} & \makecell{\textbf{Alignment}\\ \textbf{Rate(\%)}}& \textbf{Gain (\%)} & \makecell{\textbf{Selected} \\ \textbf{$k$}} & \makecell{\textbf{Alignment}\\ \textbf{Rate(\%)}}& \textbf{Gain (\%)} & \makecell{\textbf{Selected} \\ \textbf{$k$}} & \makecell{\textbf{Alignment}\\ \textbf{Rate(\%)}}& \textbf{Gain (\%)} 
\\ \hline
 					 \textbf{D1} & \textbf{k=17} & \textbf{98.95}\% & \textbf{96.3}\% & \textbf{k=17} & \textbf{98.95}\% & \textbf{96.3}\%  & \textbf{k=17} & \textbf{98.95}\% & \textbf{96.3}\%
\\ \hline
     					  \textbf{D2} & \textbf{k=17} & \textbf{61.15\%} & \textbf{80.1\%} & \textbf{k=17} & \textbf{61.15\%} & \textbf{80.1\%} & k=15 & \textbf{61.42\%} & \textbf{73.8}\% 
\\ \hline
     				 \textbf{D3} & \textbf{k=17} &  \textbf{80.39\%} & \textbf{95.34\%} & \textbf{k=15} & \textbf{80.44\%} & \textbf{86.78 \%} & \textbf{k=15} & \textbf{80.44\%} & \textbf{86.78 \%} 
\\ \hline                        
                        \textbf{D4} & \textbf{k=17} & \textbf{93.95}\% & \textbf{89.87\%} &
                         \textbf{k=17} & \textbf{93.95\%} & \textbf{89.87\%} &
                         \textbf{k=17} & \textbf{93.95\%} & \textbf{89.87\%}
\\ \hline                           
                         \textbf{D5} & \textbf{k=17} & \textbf{92.15\%} & \textbf{81.7\%} & \textbf{k = 25} & \textbf{92.09\%} & \textbf{83.8\%} &
                         \textbf{k=17} & \textbf{92.15}\% & \textbf{81.7\%} \\
  \hline
  \multicolumn{10}{|c|}{\textbf{Blue}}\\
  \hline
 					 \textbf{D1} & \textbf{k=20} & \textbf{99.53}\% & \textbf{99\%} & \textbf{k=25} & \textbf{99.29\%} & \textbf{98.6\%} & \textbf{k=20} & \textbf{99.53}\% & \textbf{99\%}
\\ \hline
     					  \textbf{D2} & \textbf{k=20} & \textbf{57.44\%} & \textbf{4.61\%} & \textbf{k=20} & \textbf{57.44\%} & \textbf{4.61\%} & \textbf{k=20} & \textbf{57.44\%} & \textbf{4.61\%} 
\\ \hline
     				 \textbf{D3}  & \textbf{ k=20} & \textbf{84.17\%} & \textbf{99.2\%} & \textbf{k=20} & \textbf{84.17\%} & \textbf{99.2\%} & \textbf{k=20} & \textbf{84.17\%} & \textbf{99.2\%} 
\\ \hline                        
                         \textbf{D4} & \textbf{ k=20} & \textbf{95.31\%} & \textbf{98.5\%} & \textbf{ k=20} & \textbf{95.31\%} & \textbf{98.5\%} & \textbf{ k=20} & \textbf{95.31\%} & \textbf{98.5\%}
\\ \hline                           
                         \textbf{D5} & \textbf{ k=20} & \textbf{92.33\%} & \textbf{88.9\%} & \textbf{ k=20} & \textbf{92.33\%} & \textbf{88.9\%} & \textbf{ k=20} & \textbf{92.33\%} & \textbf{88.9\%}\\
  \hline
  \multicolumn{10}{|c|}{\textbf{RACER}}\\
  \hline
                         \textbf{D1} & \textbf{GL=4.7M} & \textbf{99.26\%} & \textbf{84.8\%} & \textbf{GL=4.7M} & \textbf{99.26\%} & \textbf{84.8\%} & \textbf{GL=4.7M} & \textbf{99.26\%} & \textbf{84.8\%}\\ 
 \hline
     					  \textbf{D2} & \textbf{GL=4.7M} & \textbf{81.15\%} & \textbf{92.9\%} & \textbf{GL=4.7M} & \textbf{81.15\%} & \textbf{92.9\%} & \textbf{GL=4.7M} & \textbf{81.15\%} & \textbf{92.9\%}
\\ \hline
     				 \textbf{D3}  & \textbf{GL=3.7M} & \textbf{84.11\%} & \textbf{88.27\%}  & \textbf{GL=3.7M} & \textbf{84.11\%} & \textbf{88.27\%} & \textbf{GL=3.7M} & \textbf{84.11\%} & \textbf{88.27\%}
\\ \hline                        
                         \textbf{D4} & \textbf{GL=4.2M} & \textbf{95.33\%} & \textbf{97\%} & \textbf{GL=4.2M} & \textbf{95.33\%} & \textbf{97\%} & \textbf{GL=4.2M} & \textbf{95.33\%} & \textbf{97\%}
\\ \hline                           
                         \textbf{D5} & \textbf{GL=4.2M} & \textbf{92.29\%} & \textbf{81.63\%} & \textbf{GL=20M} & \textbf{92.28\%} &  \textbf{80.5\%} & \textbf{GL=4.2M} & \textbf{92.29\%} & \textbf{81.63\%}
  \\ \hline 
  \end{tabular}}
\label{tb1:Lighter-Blue-Racer-Perplexity-vs-Alignment}
\end{table}
\vspace{-8pt}

\vspace{-8pt}
\subsection{Char-RNN Language Model Results}
For training our RNN, we used the ``tensorflow-char-rnn'' library~\cite{45381}. After parameter tuning, we use the following for our experiments: 2 hidden layers with 300 neurons per layer, output layer with size 4 (i.e., corresponding to the four characters(A,C,G, and T)) mini-batch size, and learning rate of 200 and $2 e^{-3}$ respectively.
For each of the 5 datasets, we used 90\% for training, 5\% for validation, and 5\% for testing, with no overlap.

For our char-RNN results, we find that the perplexity metric has a strong negative relation to the overall alignment rate [Table \ref{tbl:Fiona_5_datasets_Correlation_N_Gram}], with the absolute value of the correlation always greater than 0.93.
Here we have to sample the uncorrected set for calculating the perplexity measure because using an RNN to perform the calculation is expensive. This is because it involves, for predicting each character, doing 4 feed-forward passes (corresponding to the one-hot encodings for A, T, G, or C), each through 600 neurons. Empirically, for a test sample size of 50K, this translates to approximately 30 minutes on a desktop-class machine.  
In the experiments with the real datasets, we use $50K$ samples with uniform sampling, and in synthetic experiments, we use $100K$ samples (i.e., only 1\% of the dataset).
We find that the perplexity score decreases roughly linearly with increasing dataset size.
More importantly though, the strong quantitative relationship between perplexity and the EC quality is maintained throughout.

\begin{table}[H]
\centering
\caption {Comparison of Overall Alignment Rate between Fiona's and RACER's (with and without \name's tuning). Columns 5 \& 6 demonstrate the strong anti-correlation values between Perplexity and Alignment Rate. The last two columns show the assembly quality (in terms of NG50) before and after correction by Racer tuned with \name. Improvements in NG50 are shown between  parentheses}
\scalebox{0.72}{
\begin{tabular}{ |c|c|c|c|c|c|c|c| }
\hline
\makecell{Dataset} & \makecell{Fiona \\ + Bowtie2 \\ (Alignment Rate)} & \makecell{RACER \\ w/o \name \\ + Bowtie2 \\ (Alignment Rate)} & \makecell{RACER \\ w/ \name \\ + Bowtie2 \\ (Alignment Rate)}& \makecell{Correlation \\ (N-Gram)} & \makecell{Correlation \\ (RNN)} & \makecell{NG50 \\ of Velvet \\ w/o EC} & \makecell{NG50 \\ of Velvet \\ w/ (Racer+\name)} \\ 
\hline
D1 &  99.25\% & 85.01\% & \textbf{99.26}\% &  -0.977 & -0.938 & 3019 & 6827 (2.26X)\\
D2 &  73.75\% & 58.66\% & \textbf{81.15}\%&  -0.981 & -0.969 & 47 & 2164 (46X) \\
D3 &  83.12\% & 80.79\% & \textbf{84.11}\% &  -0.982 & -0.968 & 1042 & 4164 (4X) \\
D4 &  95.33\% & 93.86\% &	95.33\% &  -0.946 & -0.930 & 118 & 858 (7.27X) \\
D5 &  \textbf{92.34}\% & 90.91\% &	92.29\% &  -0.970 & -0.962 & 186 & 2799 (15X) \\
\hline
\end{tabular}}
\label{tbl:Fiona_5_datasets_Correlation_N_Gram}
\end{table}

\vspace{-12pt}
\subsection{Comparison with Self-tuning EC Tool}
\vspace{-5pt}
Here, we compare \name with the EC tool, Fiona \cite{schulz2014fiona}, which estimates parameters automatically.
The purpose of this comparison is to show that \name can tune $k$-mer-based approaches (RACER specifically for this experiment) to achieve comparable performance to suffix array-based approaches (\eg Fiona), reducing the gap between the two approaches.
\cite{yang2012survey} and \cite{molnar2014correcting} show a similar comparison between different EC approaches concluding that the automatic selection of configuration parameters, based on the datasets, is crucial for EC performance.
 Table \ref{tbl:Fiona_5_datasets_Correlation_N_Gram} presents the overall alignment rate for our 5 evaluation datasets, calculated after doing correction by Fiona. We notice that RACER, when tuned with \name, outperforms automatic tuning by Fiona in 3 of the 5 datasets, while they were equal in one dataset. Finally, Fiona is only better on $D_{5} $ by $0.05\%$, which may be deemed insignificant.

\vspace{-12pt}
\subsection{Impact on Assembly Quality}
\vspace{-5pt}

Here we show the impact of tuning EC tools on genome assembly quality. We use Velvet \cite{zerbino2008velvet} to perform the assembly and QUAST \cite{gurevich2013quast} to evaluate the assembly quality. We compare the NG50 before and after correction done by RACER using the best GL found by \name. The results (Table \ref{tbl:Fiona_5_datasets_Correlation_N_Gram}) show a significant improvement on NG50 by 2.26X, 46X, 4X, 7.27X, and 15X. For D1, the improvement in NG50 before and after EC is the lowest, since D1 contains the least fraction of errors to start with. 
These improvements are consistent with what was reported in \cite{heydari2017evaluation} and \cite{greenfield2014blue}. In these prior works, improvement in genomic assembly was measured due to the use of error correction tools
with manually tuned configuration parameters.

\vspace{-5pt}
\subsection{Search time improvement with \name}
\vspace{-5pt}

\begin{table}[h]
\centering
\caption {Search time comparison for estimating the perplexity metric with \name (N-gram) for a point in search space \textit{vs}. estimating overall alignment rate with Bowtie2 }
\scalebox{0.83}{
\begin{tabular}{ |c|c|c|c|c|c|c|c|c|c| }
\hline
\multicolumn{10}{|c|}{\textbf{Dataset}}\\
\hline
 \multicolumn{2}{|c|}{\textbf{D1}} & \multicolumn{2}{|c|}{\textbf{D2}} & \multicolumn{2}{|c|}{\textbf{D3}} & \multicolumn{2}{|c|}{\textbf{D4}} & \multicolumn{2}{|c|}{\textbf{D5}}\\
  \hline
Athena & Bowtie2 & Athena & Bowtie2 & Athena & Bowtie2 & Athena & Bowtie2 & Athena & Bowtie2 \\ 
 \hline
1m 38s & 10m 5s  &  49s & 3m 53s &  1m 39s & 7m 50s &  52s & 3m 8s &  1m 40s & 9m 42s\\ 
 \hline
\end{tabular}}
\label{tbl:N_Gram_Run_Time_Vs_Bowtie}
\end{table}

Consider that in our problem statement, we are trying to search through a space of configuration parameters in order to optimize a metric (EC Gain or Alignment Rate). The search space can be large and since the cost of searching shows up as a runtime delay, it is important to reduce the time that it takes to evaluate that metric of each search point. In today's state-of-the-art, to find the best value of a configuration parameter~\cite{chikhi2013informed, mahadik2017scalable}, \eg $k$-value, the method would be to pick a $k$ (a single point in the space), run the EC tool with that value, then perform alignment (with one of several available tools such as Bowtie2), and finally compute the metric (alignment rate or EC gain) for that value. In contrast, with \name, to explore one point in the search space, we run the EC algorithm with the $k$-value, and then compute the perplexity metric, which does not involve the time consuming step of alignment. Here, we evaluate the relative time spent in exploring one point in the search space using \name vis-$\grave{a}$-vis the baseline, the state-of-the-art. The result is shown in Table \ref{tbl:N_Gram_Run_Time_Vs_Bowtie}. For this comparison, the alignment is done by Bowtie2~\cite{langmead2012fast}. We find that using the baseline approach, each step in the search takes respectively 6.2X, 4.8X, and 4.7X, 3.6X, and 5.82X for the 5 datasets respectively. Further, while we use the hill-climbing technique to search through the space, today's baseline methods use exhaustive search, such as in Lighter~\cite{song2014lighter} and thus the end-to-end runtime advantage of \name will be magnified. 

\vspace{-12pt}
\section{Related Work}
\vspace{-12pt}
\textbf{Error Correction Approaches:} EC tools can be mainly divided into three categories: $k$-spectrum based, suffix tree/array-based, and multiple sequence alignment-based (MSA) methods. Each tool takes one or more configuration parameters. While we have experimented with \name applied to the first kind, with some engineering effort, it can be applied to tuning tools that belong to the other two categories.

\noindent \textbf{Language Modeling in Genomics:} In the genomics domain, LM was used in \cite{ganapathiraju2002comparative} to find the characteristics of organisms in which N-Gram analysis was applied to 44 different bacterial and archaeal genomes and to the human genome. 
In subsequent work, they used N-Gram-based LM for extracting patterns from whole genome sequences. Others~\cite{coin2003enhanced} have used language modeling to enhance domain recognition in protein sequences. For example, \cite{king2007ngloc} has used N-Gram analysis specifically to create a Bayesian classifier to predict the localization of a protein sequence over 10 distinct eukaryotic organisms. 
RNNs can be thought of as a generalization of Hidden Markov Models (HMMs) and HMMs have been applied in several studies that seek to annotate epigenetic data. For example, \cite{song2015spectacle} presents a fast method  using spectral learning with HMMs for annotating chromatin states in the human genome.
Thus, we are seeing a steady rise in the use of ML techniques traditionally used in NLP being used to make sense of genomics data.  

\noindent \textbf{Automatic Parameter Tuning:} The field of computer systems has had several successful solutions for automatic configuration tuning of complex software systems. Our own work \cite{mahgoub2017rafiki} plus others \cite{van2017automatic} have shown how to do this for distributed databases, while other works have done this for distributed computing frameworks like Hadoop \cite{bei2016rfhoc, li2014mronline} or cloud configurations \cite{alipourfard2017cherrypick}. We take inspiration from them but our constraints and requirements are different (such as, avoiding reliance on ground truth corrected sequences). 
\vspace{-12pt}
\section{Discussion}
\vspace{-5pt}
A single iteration in \name assumes that the perplexity metric is convex in relation to the value of $k$. Intuitively, with small $k$-values, most $k$-mers will have high frequencies and hence very few will be corrected. In contrast, with high $k$-values, the number of unique $k$-mers increases, and hence no subset of $k$-mers is of high-enough fidelity. Based on the above rationale, we expect \name to perform accurately for most datasets, as hill climbing search reaches optimal solutions for convex problems. However, for non-convex spaces, a single iteration in \name may get stuck in local optima and therefore several iterations (with different intial points) is needed.  Moreover, some EC tools have a number of performance-sensitive configuration parameters, with interdependencies. For such tools, systems such as Rafiki \cite{mahgoub2017rafiki} can encode such dependencies, while relying on \name's LM to compute the corresponding performance metric, converging toward optimal parameters.

\vspace{-12pt}
\section{Conclusion}
\vspace{-6pt}
The performance of most EC tools for NGS reads is highly dependent on the proper choice of its configuration parameters, \eg $k$-value selection in \textit{k}-mer based techniques as shown in Table \ref{tb1:Lighter-Blue-Racer-Perplexity-vs-Alignment}. Using our \name suite, we target the problem of automatically tuning these parameters using language modeling techniques from the NLP domain without the need for a ground truth genome. 
Through N-Gram and char-RNN language modeling, we validate the intuition that the EC performance can be computed quantitatively using the ``perplexity'' metric, which then guides a hill climbing-based search toward the best $k$-value. 
We evaluate \name with 5 different real datasets, plus, with synthetically injected errors. We find that the predictive performance of the perplexity metric is maintained under all scenarios, with absolute correlation scores higher than 0.93. Further, using the perplexity metric, \name can search for and arrive at the best $k$-value, or within 0.27\% of the assembly quality obtained using brute force.


\bibliographystyle{myrecomb}

\bibliography{athena}

\section{Appendix}
\vspace{-5pt}
\subsection{Background}
\label{sec:Background}
\noindent{\bf Error Correction and Evaluation} 

The majority of error correction tools share the following intuition: high-fidelity sequences (or, solid sequences) can be used to correct errors in low-fidelity sequences (or, in-solid sequences). However, they vary significantly in the way they differentiate between solid and in-solid sequences. For example, \cite{yang2010reptile} corrects genomic reads containing insolid $k$-mers using a minimum number of edit operations such that these reads contain only solid $k$-mers after correction.
The evaluation of \textit{de novo} sequencing techniques rely on likelihood-based metrics such as ALE and CGAL, without relying on the availability of a reference genome. On the other hand, comparative sequencing or re-sequencing, such as to study structural variations among two genomes, do have reference genomes available. 

\noindent{\bf Language Modeling}

To increase the accuracy of detecting words in speech recognition, language modeling techniques have been used to see which word combinations have higher likelihood of occurrence than others, thus improving context-based semantics. Thus, language modeling is being used in many applications such as speech recognition, text retrieval, and many NLP applications. The main task of these statistical models is to capture historical information and predict the future sequences based on that information.

\noindent{\bf N-Gram-Based Language Modeling}. 
This type of modeling is word-based.
The main task that N-Gram based models \cite{brown1992class} have been used for is to estimate the likelihood of observing a word \textit{$W_i$}, given the set of previous words ${W_0, \cdots W_{i-1}}$, estimated using the following equation:
\newline
\begin{equation}
\begin{aligned}
  P(W_0, W_1, ..., W_{m}) &= \prod_{i=1}^{m}  P(W_{i} |  W_{i-1}, ..., W_{1}) \\
  &\approx \prod_{i=1}^{m}  P(W_{i} |  W_{i-1}, ..., W_{i-n})
  \end{aligned}
\end{equation}
\begin{figure}
  \includegraphics[width=0.9\linewidth]{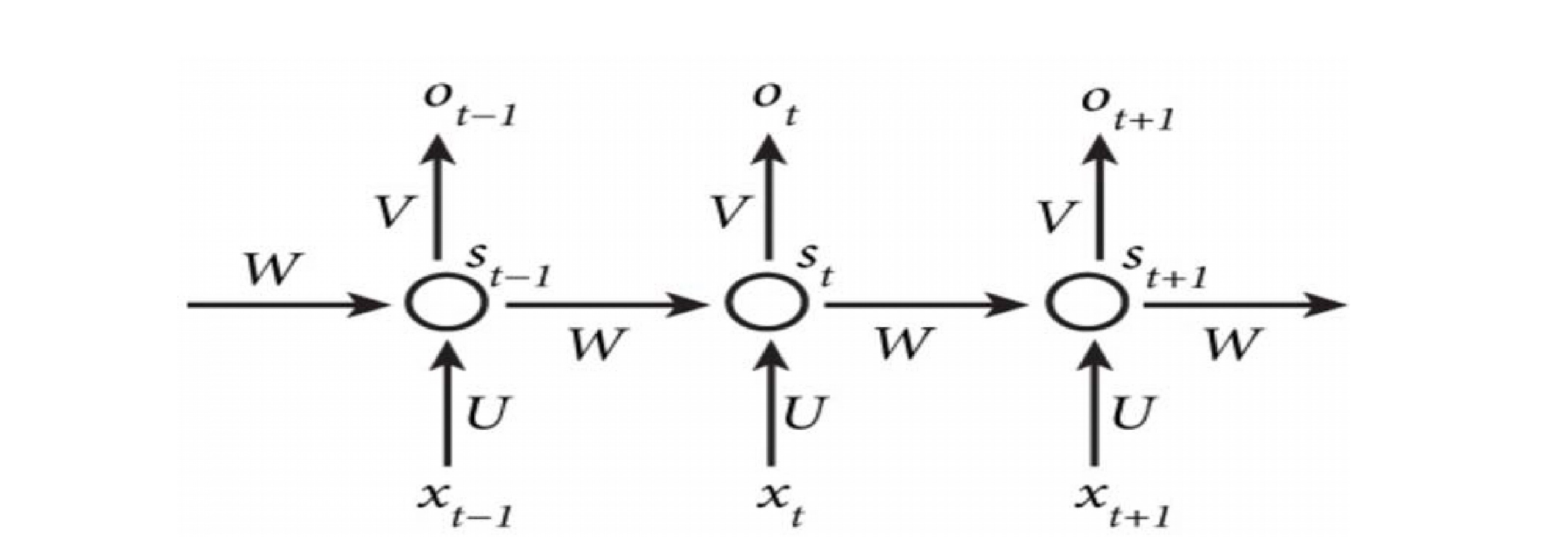}
  \caption{Structure of a Recurrent Neural Network consisting of a chain of repeating modules of neural networks. The RNN predicts the next character at each time step $(t+1)$, knowing the history of characters in previous time steps.}
\label{fig:RNN}
\end{figure}
\noindent where \textit{n} represents the number of history words the model uses to predict the next word. Obviously, a higher $n$ results in better prediction, at the cost of higher training time resulting from a more complex model.  

\noindent{\bf Char-RNN-Based Language Modeling}.
Recurrent neural network (RNN) is a very popular class of neural networks for dealing with sequential data, frequently encountered in the NLP domain. The power of RNN is that each neuron or unit can use its internal state memory to save information from the previous input and use that state, together with the current input, to determine what the next output should be. 
Character-level RNN models, \textit{char-RNN} for short, operate by taking a chunk of text and modeling the probability distribution of the next character in the sequence, given a sequence of previous characters. This then allows it to generate new text, one character at a time~\cite{graves2013generating}.
As shown in Fig. \ref{fig:RNN}, RNNs consist of three main layers: Input Layer, Hidden Layer, and Output Layer.
First, Input Layer takes $ x_{t} $ vector, which is input at a time step $t$, usually a one-hot encoding vector of the $ t^{th} $ word or character of the input sentence. 
Second, Hidden Layer consists of the hidden state at the same time step $ s_{t} $, which represents the memory of this network. It is calculated as a non-linear function \textit{f} (\eg tanh) of the previous hidden state $ s_{t-1} $ and the input at current time step $ x_{t} $ with the following relation: \begin{equation}
s_{t} = f(U x_{t} + W s_{t-1}).
\end{equation}
Here, $W$ is a matrix that consists of hidden weights of this hidden layer.
Finally, Output Layer consists of a vector $ o_{t} $, which represents the output at step $t$ and contains prediction probabilities for the next character in the sentence. Formally, its length equals the size of the vocabulary and is calculated using a softmax function.
Backpropagation was used to train the RNN to update weights and minimize the error between the observed and the estimated next word. For Deep RNN architectures, there are multiple parameters that affect the performance of the model. The two main parameters are: \textit{Number of Hidden Layers} and \textit{Number of Neurons per Layer}. For Char-RNN language modeling, vocabulary would include the four nucleotide bases as characters {A, C, G, and T}. Each input is a one-hot encoding vector for the four nucleotides.
Each output vector at each time step also has the same dimension.  

\noindent{\bf Perplexity of the Language Model}. 
Perplexity is a measurement of how well a probability distribution predicts a sample. In NLP, perplexity is one of the most effective ways of evaluating the goodness of fit of a language model since a language model is a probability distribution over entire sentences of text~\cite{azzopardi2003investigating}. For example, 5 per word perplexity of a model translates to the model being as confused on test data as if it had to select uniformly and independently from 5 possibilities for each word. Thus, a lower perplexity indicates that language model is better at making predictions.

\subsection{Detailed Results}

In this section, we show a detailed version of the results presented in Table \ref{tb1:Lighter-Blue-Racer-Perplexity-vs-Alignment}. 
\begin{table}
\centering
\small
\begin{tabular}{ |c|c|c|c|c|c|c| } 
\hline
EC Tool & {Dataset} & Tuned Parameter & \makecell{Perplexity\\(RNN)} & \makecell{Perplexity\\(N-Gram)} & \makecell{Overall Alignment\\ Rate} & Gain (\%) \\  
 \hline
 					  &  & k = 10 & 103.00579 & 20.42 & 97.45\% & -0.01\% \\ 
 					  & D1 & k = 15 & 103.0048 & 16.86 & 98.83\% & 87.5\% \\ 
 					  &  & \textbf{k = 17} & \textbf{103.004} &  \textbf{16.7}  & \textbf{98.95}\% & \textbf{96.3}\% \\ 
 					  &  & k = 25 &  103.00551 &  18.22 & 97.98\% & 69.5\%  
\\\cline{2-7}
     					  &  & k = 10 & 204.849 & 121.88 & 56.9\% & 0\%  \\ 
						  & D2 & \textbf{k = 15} & 204.775 & 102.13 & \textbf{61.42\%} & 73.8\% \\ 
     					  &  & k = 17 &  \textbf{204.76} &  \textbf{100.30}  & 61.15\% & \textbf{80.1}\% \\
     					  &  & k = 25 & 204.795 & 107.37 & 59.19\%  & 69.96\%
 \\\cline{2-7}
     		\multirow{4}{*}{Lighter} &  & k = 10 & 200.513 & 52.81 & 72.91\% & 0\% \\ 
     					  & D3 & \textbf{k = 15} & 200.432 & 33.2602 & \textbf{80.44\%} & 86.78 \% \\ 
     					  &  & k = 17 & \textbf{200.432} & \textbf{32.74}  & 80.39\% & \textbf{95.34\%}\\ 
     					  &  & k = 25 & 200.529 & 42.28 & 75.33\% & 65\%
 \\\cline{2-7}                         
                           &  & k = 10 & 207.2945 & 25.53 & 92.14\% & 0\%\\
                           & D4 & k = 15 & 206.248 & 18.07 & 93.72\% & 86.14 \%\\
                           &  & \textbf{k = 17} & \textbf{204.899} & \textbf{17.86} & \textbf{93.95}\% & 89.87\%\\
                             &  & k = 25 & 206.848 & 18.25 &  93.13\% & \textbf{89.9} \%
\\\cline{2-7}                            
                            &  & k = 10 & 193.121 & 6.44 & 91.92\% & 0\%\\
                            & D5 & k = 15 & 193.052 & 5.45 & 92.11\% & 73.12\% \\
                            &  & \textbf{k = 17} & 193.054  &  \textbf{5.354}  & \textbf{92.15}\% & 81.7\% \\
                 &  & k = 25 & \textbf{193.052} & 5.357 & 92.09\% & \textbf{83.8}\%\\                 
  \hline
  \end{tabular}
\caption {Detailed results for our 5 datasets using Lighter: a comparison between finding best value of $k$ using \name variants vs exhaustive seraching. These results are consistent with the reported results by Lighter's authors (see Figure 5 in \cite{song2014lighter}).}
\label{tb1:Lighter-Perplexity-vs-Alignment}
\end{table}


\begin{table}
\centering
\small
\begin{tabular}{ |c|c|c|c|c|c|c| } 
\hline
EC Tool & {Dataset} & Tuned Parameter & \makecell{Perplexity\\(RNN)} & \makecell{Perplexity\\(N-Gram)} & \makecell{Overall Alignment\\ Rate} & Gain (\%) \\ 
 \hline
						 &  & \textbf{k = 20} &  206.033 &  \textbf{16.52} & \textbf{99.53}\% & \textbf{99\%} \\ 
					   & D1 & k = 25 &  \textbf{206.026} &  16.62 & 99.29\% & 98.6\%\\ 
					   &  & k = 30 &  206.0361 &  16.96 & 98.65\% & 87.6\% 
 \\\cline{2-7}
 \multirow{4}{*}{Blue} &  & \textbf{k = 20} & \textbf{204.846} & \textbf{119.1738} & \textbf{57.44\%} & \textbf{4.61\%} \\ 
 					   & D2 & k = 25 &  204.848 & 120.5232 & 57.09\% & 1.7\% \\ 
 					   &  & k = 30 & 204.847 & 238.98 & 57\%  & 1.24\%  
 \\\cline{2-7}
 					  &  & \textbf{k = 20} & \textbf{200.46}  & \textbf{29.89} & \textbf{84.17}\% & \textbf{99.2\%} \\ 
  					  & D3 & k = 25 & 200.49   & 32.39 & 81.62\% & 97.7\% \\ 
  					  &  & k = 30 & 200.51 & 49.22 & 73.84\% & 13.17 \%
 \\\cline{2-7}                     
                        &  & \textbf{k = 20} & \textbf{207.179} & \textbf{46.6} & \textbf{95.31\%} & \textbf{98.5\%} \\
                       & D4 & k = 25 & 207.228 & 47.59 & 94.64\% & 98.4\% \\  
                &  & k = 30 & 207.284 & 48.69 & 93.97\%   & 96.5\%     
                         
\\\cline{2-7}
                &  & \textbf{k = 20} & \textbf{192.804} & \textbf{15.67} & \textbf{92.33}\% & 88.9\% \\   
               & D5 & k = 25 & 192.8044 & 15.72 & 92.28\% & \textbf{91.2\%} 
\\               
                &  & k = 30 & 192.8077 & 15.79 & 92.22\%  & 92.08\%
\\           
  \hline
  \end{tabular}
\caption {Detailed results for our 5 datasets using Blue}
\label{tb1:Blue-Perplexity-vs-Alignment}
\end{table}


\begin{table}
\centering
\small
\begin{tabular}{ |c|c|c|c|c|c|c| } 
\hline
EC Tool & {Dataset} & Tuned Parameter & \makecell{Perplexity\\(RNN)} & \makecell{Perplexity\\(N-Gram)} & \makecell{Overall Alignment\\ Rate} & Gain(\%) \\ 
 \hline
 						 & D1 & \textbf{GL = 4.7M} & \textbf{206.033} & \textbf{16.6} & \textbf{99.26}\% & \textbf{84.8\%} 
 \\ 						  
						 &  & GL = 20M & 206.036 & 16.9 &  98.85\% &  80.3\% \\ 
 						 &  & GL = 30M & 206.0357 & 16.99 & 98.82\% & 77.6\%
 \\\cline{2-7}
  						 & D2 & \textbf{GL = 4.7M} & \textbf{204.7520} & \textbf{85.14}  & \textbf{81.15}\% & 92.9\% \\
                         &  & GL = 7M & 204.7564 & 85.2  & 81.13\% & \textbf{93\%}\\
                         &  & GL = 30M & 204.775 & 88.24  & 79.24\% & 91.9\%
  \\\cline{2-7}                       
               \multirow{4}{*}{Racer}          &  & \textbf{GL = 3.7M} & 200.4552 & \textbf{30.4}  & \textbf{84.11}\% & 88.27\% \\                                               
                         & D3 & GL = 18M & 200.4603 & 33.87 & 80.97\% & 80.21\% \\
                         &  & GL = 30M & 200.4626 & 34.46  & 80.79\%b & 75.74\%
  \\\cline{2-7}                       
                         & D4 & \textbf{GL = 4.2M} & \textbf{206.942} & \textbf{17.32} & \textbf{95.33}\% & \textbf{97\%} \\
                         &  & GL = 20M & 206.9494 & 17.51 & 95.04\% & 96.5\% \\
                         &  & GL = 30M & 206.9489 & 17.53 & 95.01\% & 95.9\%
  \\\cline{2-7}                       
                         
                         & D5 & \textbf{GL = 4.2M} & 193.0454 & \textbf{4.77} & \textbf{92.29}\% & 81.63\%\\
                         &  & GL = 20M & \textbf{193.0451} & 4.78  & 92.28\% & 80.5\% \\
                         &  & GL = 30M & 193.0479 & 4.79  & 92.26\% & 81.9\% \\
  \hline
  \end{tabular}
\caption {Detailed results for our 5 datasets using Racer.}
\label{tb1:Racer-Perplexity-vs-Alignment}
\end{table}

\subsection{Evaluation with Synthetically Injected Errors}
Here we experiment on data sets where we synthetically inject errors of three kinds - insertion, deletion, and substitution. The real data sets used in our evaluation in Section \ref{sec:evaluation-real} belonged to Illumina and therefore had primarily substitution errors (about 99\% of all errors). Hence our synthetic injections are meant to uncover if the relationship between perplexity and error rate holds for the other error types. 
We start by randomly collecting 100K short reads from the reference genome (\ie, without any error) for each of the two organisms used in the real data sets---\textit{E. coli} (D1, D2) and \textit{Acinetobacter} (D3). Afterward, we inject errors of each type as follows:
\begin{enumerate}[leftmargin=*]
\item \textbf{Deletion:} We select the index of injection (position from which to delete sequence), which follows a uniform distribution U(0,$l-d$), where $l$ is the length of the read and $d$ is the length to delete.
\item \textbf{Insertion:} Similar to deletion errors, the index of insertion follows a uniform distribution U(0,$l-I$), where $l$ is the length of the read and $I$ is the length of inserted segment. Moreover, each base pair in the inserted segment follows a uniform distribution over the four base pairs (A, C, G, or T).
\item \textbf{Substitution:} For this type of synthetic errors, we select $k$ positions of substitutions following a uniform distribution U(0,$l$), where $l$ denotes the length of the read. Second, we substitute each base pair of the $k$ positions with another base pair following a uniform distribution over the four base pairs (A, C, G, or T). Note that there is one-in-four chance that the substitution results in the same base pair. 
\end{enumerate}
We test the proposed approach against two levels of synthetic errors: low error rate (Uniform in (1, 5) bp in error per read), and high error rate (Uniform in (6, 10) bp in error per read). In all cases, the language model is trained using the original data set, mentioned in Table \ref{tbl:Datasets_Description}, with the natural ambient rate of error. 
Figure \ref{figure:Synthetic_Errors_Detection} shows the results for the perplexity metric in the data set {\em after} error injection (\ie, without doing correction), for the three error types and an equi-probable mixture of the three. The results show that for all types of errors, the direct quantitative relation between error rate and perplexity holds---the perplexity values are higher for high error rate injection. One observation is that the values of perplexity for insertion and substitution errors are higher than for deletion errors. For example, insertion and substitution perplexities are 3X \& 4.7X the deletion perplexity for high error rate data sets, and 2X \& 1.5X for low error rate data sets. This is expected because in deletion errors, the segment of the read after the index of deletion is a correct segment and hence is expected by the language model. On the other hand, for insertion and substitutions, the added synthetic base pairs will most likely construct wrong segments. Such segments will have very low counts in the language model and therefore produce higher overall perplexities.
Another observation is that the perplexities for the injection into data set D1 are higher than for D3. This is likely because D3 had a higher natural ambient rate of error and hence the additional injection does not cause as much increase in the perplexity metric.

\begin{figure}[t] \label{figure:Synthetic_Errors_Detection}
\begin{minipage}[t]{1.0\textwidth}
\begin{minipage}[t]{.24\textwidth}
\centering
  \includegraphics[width=\linewidth]{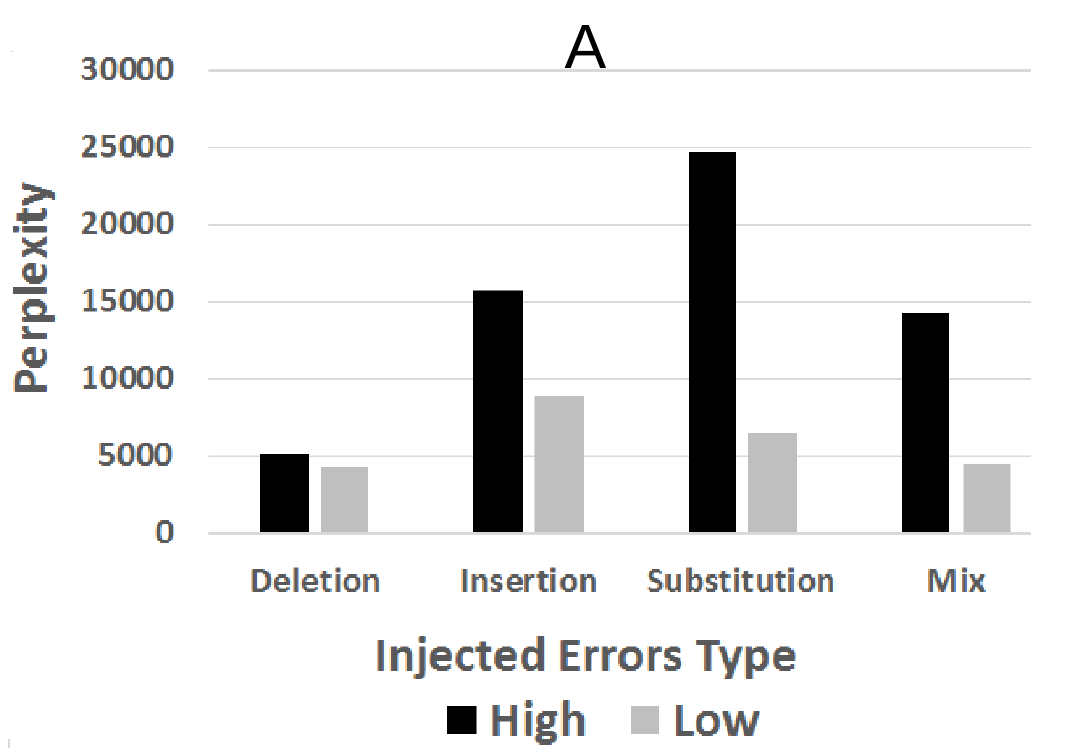}
\end{minipage}%
\begin{minipage}[t]{.24\textwidth}
\centering
\includegraphics[width=\linewidth]{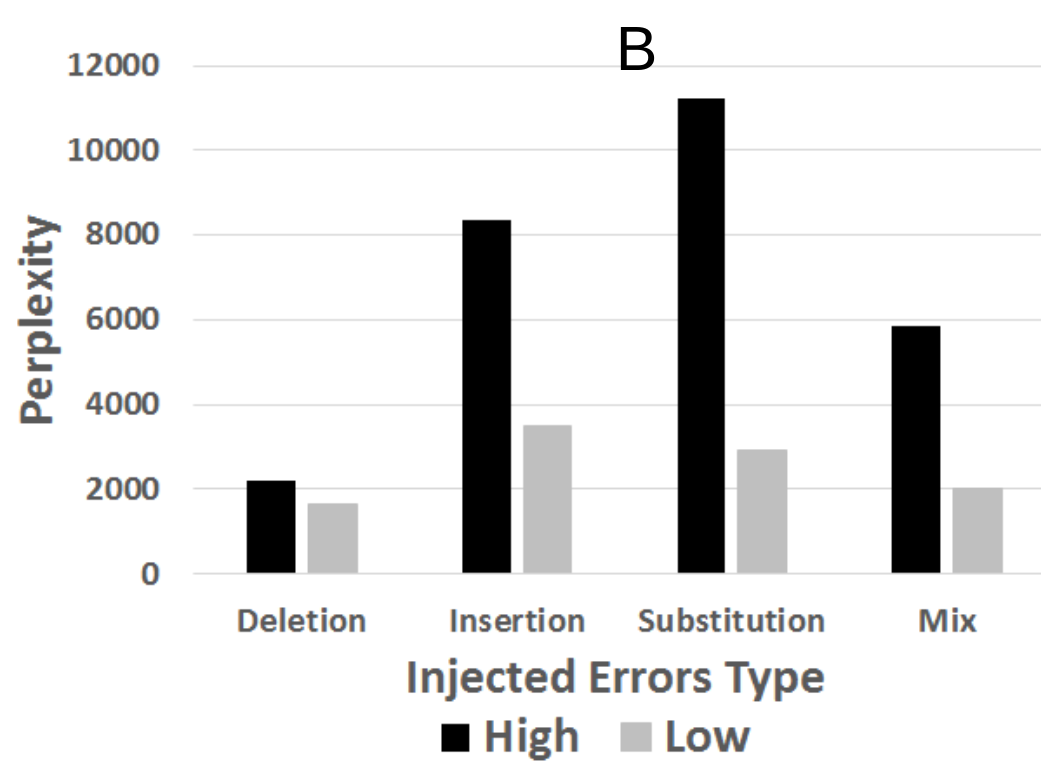}
\end{minipage}%
\begin{minipage}[t]{.24\textwidth}
\centering
  \includegraphics[width=\linewidth]{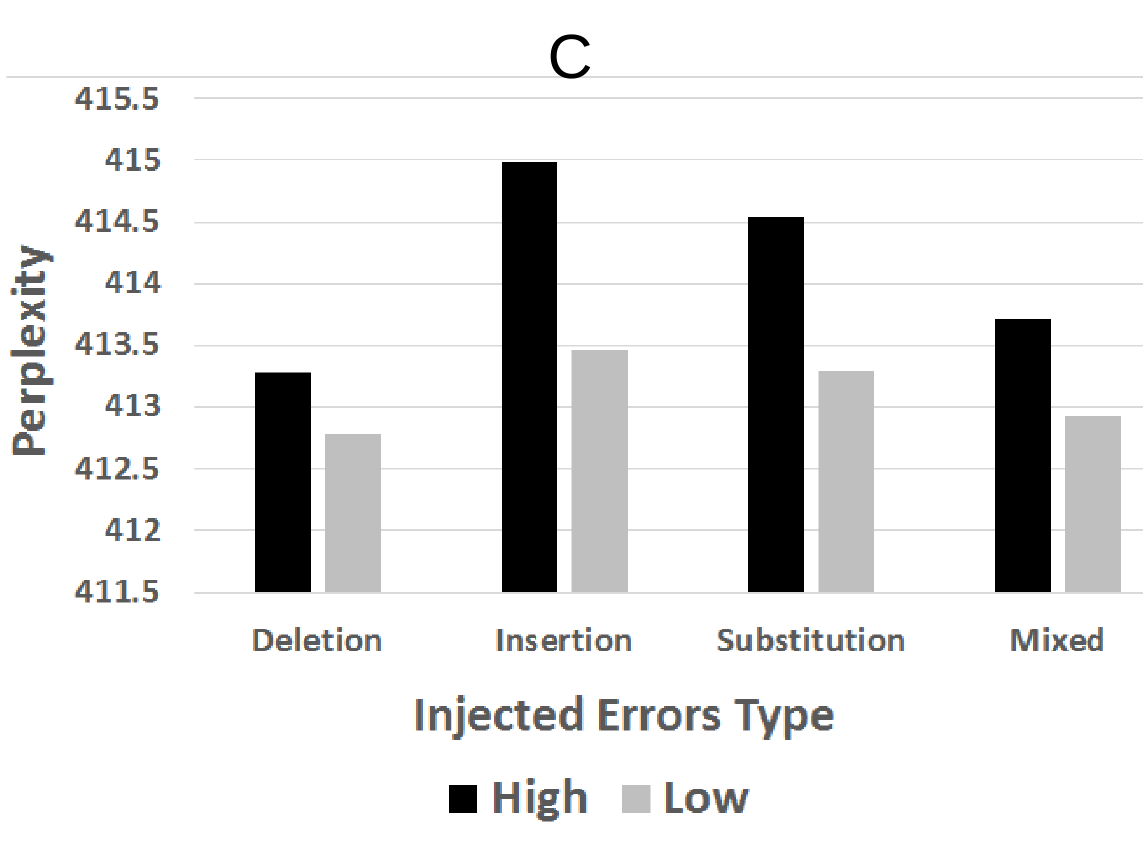}
\end{minipage}
\begin{minipage}[t]{.24\textwidth}
\centering
  \includegraphics[width=\linewidth]{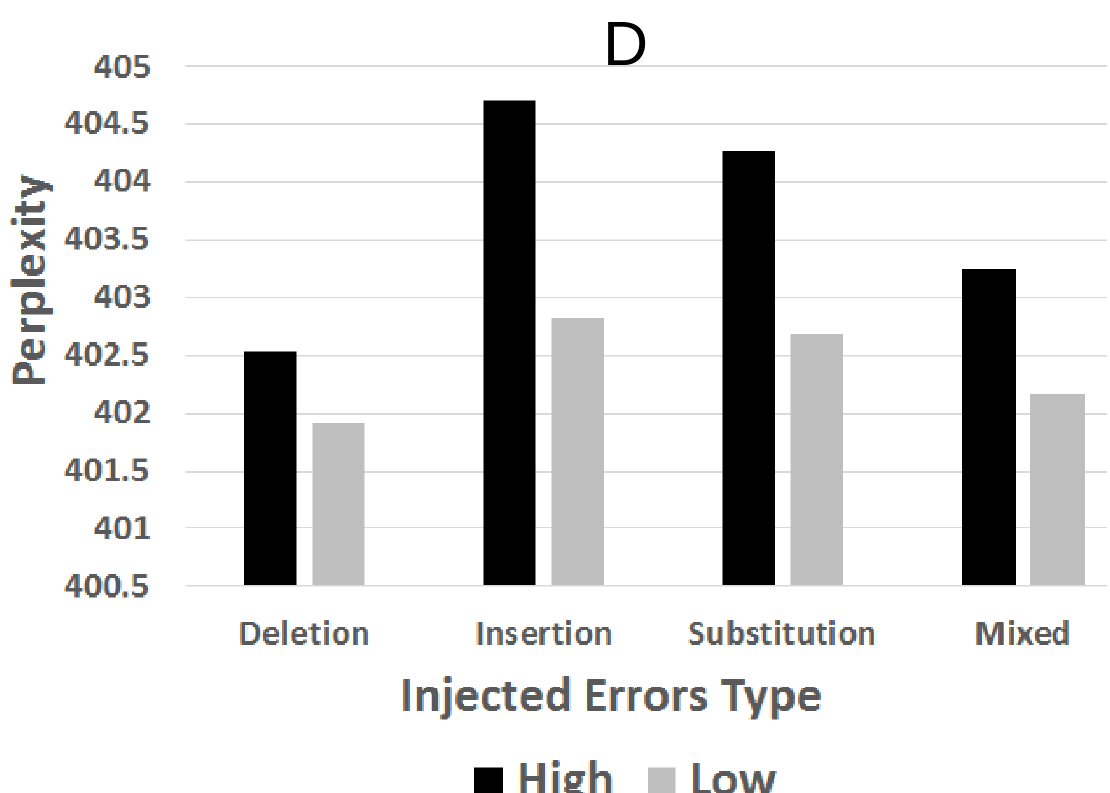}
\end{minipage}
\end{minipage}
  \caption{N-Gram (Figures A and B) and RNN (Figures C and D)  Perplexity metric for different types of synthetic errors: Indels and Substitution errors, and a mixture of the three for \textit{E. coli} str reference genome (Figures A and C) and \textit{Acinetobacter} sp. reference genome (Figures B and D). We compare two versions of such errors: high and low error rates.}
\label{figure:Synthetic_Errors_Detection}
\end{figure}


  

\end{document}